\documentclass[runningheads,a4paper]{llncs}
\usepackage[top=2.5cm, bottom=2.5cm, left=2.5cm, right=2.5cm]{geometry}
\usepackage{amsmath}
\usepackage{amssymb}
\usepackage{graphicx}
\usepackage{booktabs}
\usepackage{listings}
\usepackage{xcolor}
\usepackage{hyperref}
\usepackage{caption}
\usepackage{subcaption}

\usepackage{url}
\usepackage{color}
\usepackage{cite}
\usepackage{epsfig}

\usepackage[nomargin,inline,marginclue,draft]{fixme}
\usepackage{cleveref}
\fxsetup{status=draft}
\fxsetup{theme=color, mode=multiuser} \FXRegisterAuthor{me}{ame}{\color{red}Me}
\newcommand{\orcid}[1]{\href{https://orcid.org/#1}{\includegraphics[scale=0.02]{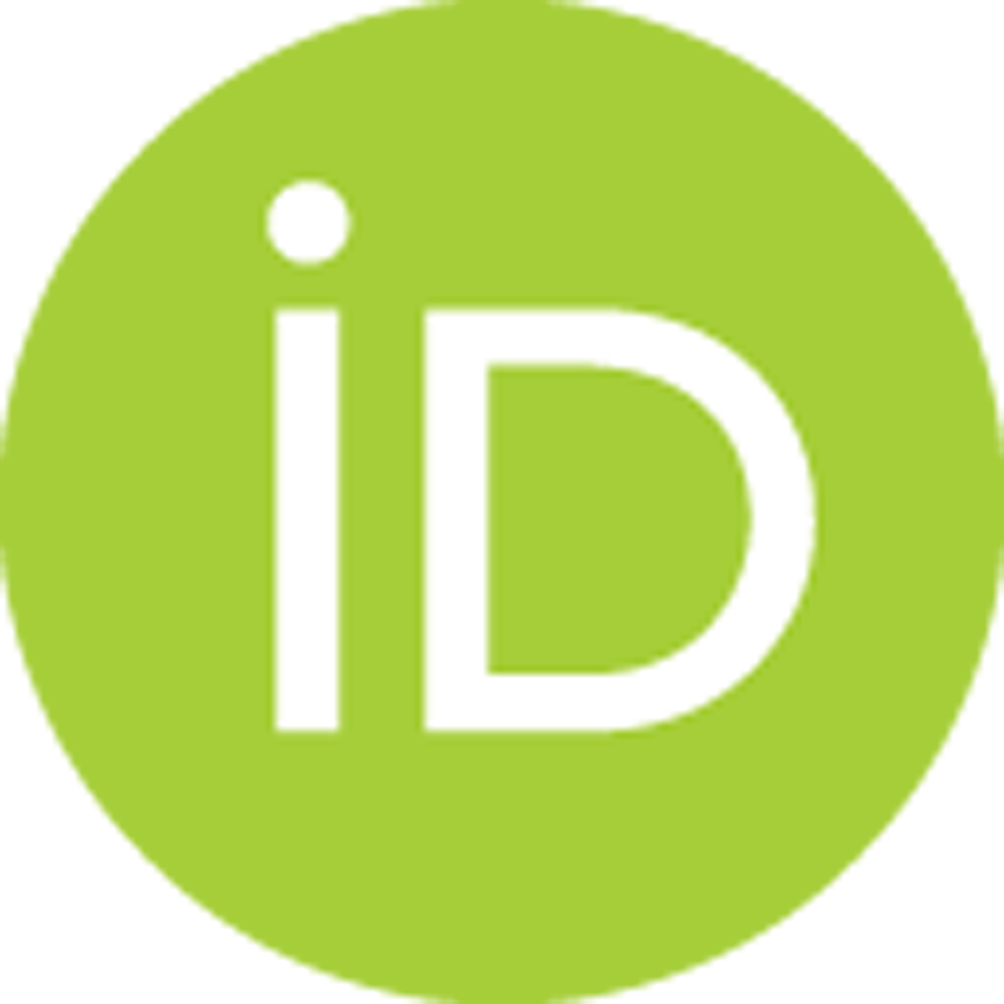}}} 

\hypersetup{
    colorlinks=true
}
\title{The Brain Resection Multimodal Image Registration (ReMIND2Reg) 2025 Challenge}
\titlerunning{The ReMIND2Reg 2025 Challenge}

\author{
Reuben Dorent \inst{1,2,3,\dag,\ddag,\S,**,\orcid{0000-0002-7530-0644}}
\and Laura Rigolo \inst{3,\ddag,\orcid{0000-0003-2519-6153}}
\and Colin P. Galvin \inst{3,\ddag}
\and Junyu Chen \inst{4,\dag,,\orcid{0000-0003-4672-6408}}
\and Mattias P Heinrich \inst{5,\dag,\orcid{0000-0002-7489-1972}}
\and Aaron Carass \inst{6,\dag,\orcid{0000-0003-4939-5085}}
\and Olivier Colliot  \inst{2,\dag,\orcid{0000-0002-9836-654X}}
\and Demian Wassermann \inst{1,\dag,\orcid{0000-0001-5194-6056}}
\and Alexandra Golby  \inst{3,\dag,\ddag, \orcid{0000-0001-8461-9561}}
\and Tina Kapur \inst{3,\dag,\ddag,\orcid{ 0000-0003-3646-9508}}
\and William Wells \inst{3,7,\dag,\ddag,\S, \orcid{0000-0002-4279-8634}}
}

\authorrunning{Dorent R, et al.}

\institute{\scriptsize{ Inria Saclay Île-de-France, CEA, Université Paris-Saclay, Palaiseau, France
\and Sorbonne Université, Institut du Cerveau - Paris Brain Institute - ICM, CNRS, Inria, Inserm, AP-HP, Hôpital de la Pitié Salpêtrière, F-75013, Paris, France
\and Harvard Medical School, Brigham and Women's Hospital, Boston, MA, USA
\and Department of Radiology and Radiological Science, Johns Hopkins Medical School, Baltimore, MD, USA
\and Institute of Medical Informatics, University of Lübeck, Lübeck, Schleswig-Holstein, Germany.
\and Image Analysis and Communications Laboratory in the Department of Electrical and Computer Engineering, Johns Hopkins University, Baltimore, MD, USA.
\and CSAIL, MIT, Cambridge MA, USA
}
\linebreak
\\
\textsuperscript{\dag} People involved in the organization of the challenge.\\
\textsuperscript{\ddag} People contributing data from their institutions.\\
\textsuperscript{\S} People involved in annotation process.\\ 
\textsuperscript{**} Corresponding author: \email{\{reuben.dorent@inria.fr\}}}

\begin{document}
    \mainmatter
    \maketitle
    \setcounter{footnote}{0} 
     \begin{abstract}
Accurate intraoperative image guidance is critical for achieving maximal safe resection in brain tumor surgery, yet neuronavigation systems based on preoperative MRI lose accuracy during the procedure due to brain shift. Aligning post-resection intraoperative ultrasound (iUS) with preoperative MRI can restore spatial accuracy by estimating brain shift deformations, but it remains a challenging problem given the large anatomical and topological changes and substantial modality intensity gap. The ReMIND2Reg 2025 Challenge provides the largest public benchmark for this task, built upon the ReMIND dataset. It offers 99 training cases, 5 validation cases, and 10 private test cases comprising paired 3D ceT1 MRI, T2 MRI, and post-resection 3D iUS volumes. Data are provided without annotations for training, while validation and test performance are evaluated on manually annotated anatomical landmarks. Metrics include target registration error (TRE), robustness to worst-case landmark misalignment (TRE30), and runtime. By establishing a standardized evaluation framework for this clinically critical and technically complex problem, ReMIND2Reg aims to accelerate the development of robust, generalizable, and clinically deployable multimodal registration algorithms for image-guided neurosurgery.
    \end{abstract}
    
    \keywords{ReMIND2Reg, ReMIND, challenge, brain, tumor, surgery, registration, ultrasound, MRI}
    
    \section{Introduction}

Surgical brain resection remains the primary treatment for most malignant brain tumors. While brain tumors encompass a heterogeneous group of neoplasms, gliomas account for approximately $25\%$ of all primary brain tumors and $80\%$ of malignant brain and central nervous system tumors in adults, according to the Central Brain Tumor Registry of the United States~\cite{price2024cbtrus}. For resectable brain tumors, achieving \textit{maximal safe resection}, i.e., removing as much tumor tissue as possible without inducing new neurological deficits,  is the most significant modifiable factor on patient survival, recurrence rates, and neurological outcomes~\cite{li2016influence}. Achieving this goal requires accurate intra-operative visualization of tumor boundaries and surrounding critical functional structures~\cite{bonosi2023maximal}.

To facilitate maximal safe resection, intra-operative imaging modalities are increasingly integrated into neurosurgical workflows. Intra-operative magnetic resonance imaging (iMRI) provides high-contrast anatomical detail, enabling precise identification of residual tumor tissue. However, iMRI systems are costly, require specialized operating suites, and interrupt surgical workflow. Intra-operative ultrasound (iUS) has emerged as a practical alternative: it is portable, low cost, and capable of generating volumetric images in minutes without significant operative disruptions. Despite these advantages, iUS presents challenges due to lower soft-tissue contrast, speckle noise, and operator dependency, making interpretation difficult without advanced image processing methods~\cite{canalini2019segmentation,dorent2024patient,carton2020automatic, dorent2023unified,behboodi2022resect, dorent2025unified}.

Neuronavigation systems have also become a standard component of modern neurosurgery, enabling real-time tracking of surgical instruments relative to preoperative MRI. While valuable early in the procedure, neuronavigation accuracy degrades as surgery progresses due to \textit{brain shift}\cite{bastos2021challenges,dorward1998postimaging,nabavi2001serial,nimsky2000quantification,orringer2012neuronavigation}, i.e., nonlinear deformations of the brain caused by cerebrospinal fluid loss, gravity, swelling, and tumor removal. This progressive misalignment often leaves the surgeon without reliable image guidance precisely when the risk of neurological injury is highest, i.e., at the boundary of the tumor.

A promising approach to compensate for brain shift is multimodal image registration. In particular, aligning intra-operative 3D iUS with preoperative MRI has the potential to restore spatial accuracy in neuronavigation systems. However, this task is particularly challenging in the post-resection setting, where large and non-linear deformations and topological changes from tissue removal complicate image registration. Further complexity arises from the large modality gap between iUS and MRI, since they significantly differ in the information they capture (morphological versus echo-based) and in resolution and noise characteristics [5]. Finally, algorithms should handle incomplete preoperative MRI data to be used in clinical practice. Indeed, not all patients have both 3D contrast-enhanced T1-weighted (ceT1) and 3D T2-weighted (T2) sequences available. Therefore, algorithms for this problem must be robust to missing modalities, adaptable to large anatomical and topological changes, and fast and accurate enough for intra-operative deployment.

The ReMIND2Reg 2025 challenge aims to encourage algorithmic solutions that address these needs. Unlike previous challenges that focused on pre-dural opening registration problems (i.e., before tissue resection and without large deformations)~\cite{hering2023tmi}, ReMIND2Reg offers the largest publicly available benchmark for registering preoperative MRI to post-resection iUS images. Building on the ReMIND dataset~\cite{remind}, the challenge provides 99 training cases (93 ceT1, 62 T2, and 99 iUS volumes) without annotations, 5 validation cases (5 ceT1, 5 T2, and 5 iUS volumes) with private annotations, and 10 private annotated test cases (10 ceT1, 10 T2, and 10 iUS volumes). Evaluation is based on target registration error (TRE) and robustness to TRE outliers.

By creating a standardized benchmark for this clinically relevant and technically challenging problem, ReMIND2Reg aims to accelerate the development of robust, generalizable, and clinically deployable registration methods that can improve surgical guidance and, ultimately, patient outcomes.
         
    \section{Materials \& Methods}
            \begin{figure}[t]
          \centering
          \includegraphics[width=0.9\linewidth]{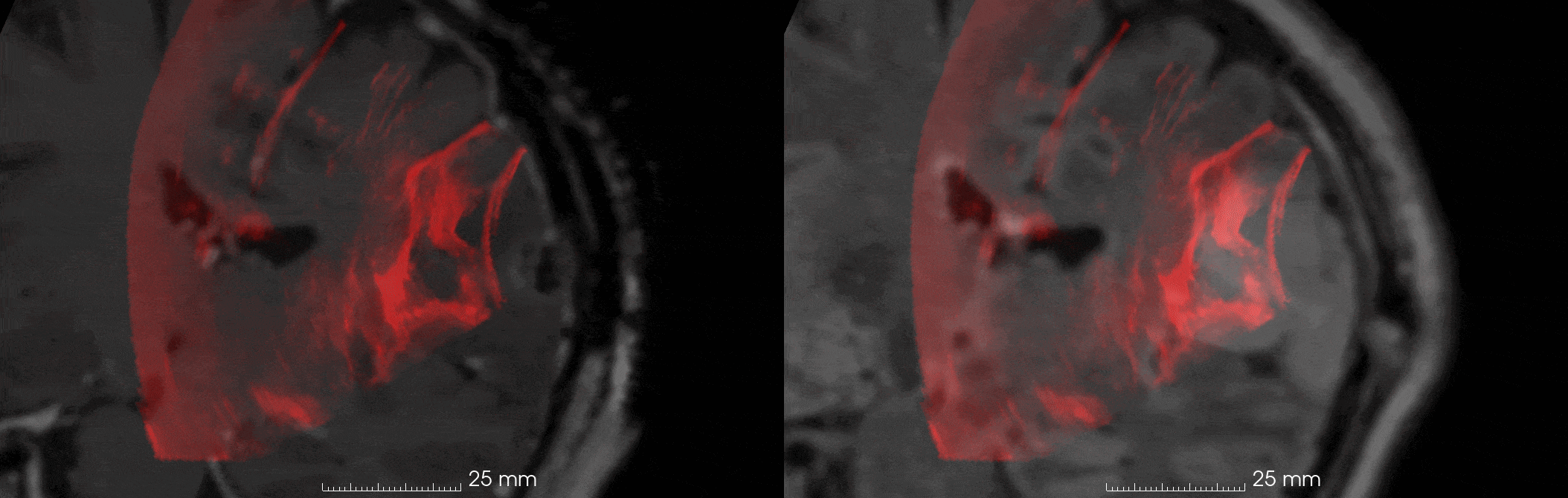}  
          \caption{\textbf{Illustrative example of unregistered preoperative MRI and overlaid post-resection iUS.} Misalignment is observed between (a) contrast-enhanced T1 MRI and post-resection iUS, and (b) T2 MRI and post-resection iUS. A prominent resection cavity is clearly visible in the iUS images.}
        \label{annotations}
    \end{figure}

\subsection{ReMIND2Reg 2025 dataset}

\label{sec:data}

    \subsubsection{Imaging Data Description.}  
    The ReMIND2Reg dataset is a pre-processed subset of the ReMIND dataset~\cite{remind}, which contains pre- and intra-operative data collected on consecutive patients who were surgically treated with image-guided tumor resection between 2018 and 2024 at the Brigham and Women’s Hospital (Boston, USA). The training (N=99) and validation (N=5) cases correspond to a subset of the public version of the ReMIND dataset. Specifically, the training set includes images of 99 patients with 99 3D iUS, 93 ceT1, and 62 T2-SPACE, and validation images of 5 patients with 5 3D iUS, 5 ceT1, and 5 T2-SPACE. The images are paired as described above, with one or two MR-iUS pairs per patient, resulting in 155 image pairs for training and 10 image pairs for validation. The 10 test cases corresponding to 20 image pairs (10 3D US, 10 ceT1, and 10 T2-SPACE) are not publicly available and remain private. 

    Collection, analysis, and release of the ReMIND database have been performed in compliance with all relevant ethical regulations. The Institutional Review Board at the Brigham and Women’s Hospital approved the protocol (2002-P-001238), and informed consent was obtained from all participants, including for public data sharing.

    \subsubsection{Acquisition details.}

    Preoperative MR data were acquired at multiple institutions in the USA, using scanners from different manufacturers. Intra-operative ultrasound (iUS) data were acquired exclusively in the Advanced Multimodality Image Guided Operating (AMIGO) suite at Brigham and Women's Hospital, Boston, MA, USA.

    The preoperative MR data includes two 3D sequences: contrast-enhanced T1-weighted (ceT1) and T2-weighted scans, though not all sequences are available for every patient. ceT1 images were obtained using MP-RAGE sequences from different vendors. T2-weighted images were acquired with the SPACE protocol on Siemens 3T systems.

    All iUS series were acquired using a sterilizable 2D neuro-cranial curvilinear transducer on a cart-based ultrasound system (N13C5, BK5000, GE Healthcare, Peabody, MA, USA) in the AMIGO suite. The ultrasound probe had a contact area of $29 \text{mm} \times 10 \text{mm}$ and a frequency range of 5-13 MHz. The imaging plane was chosen to be as parallel as possible to one of the three cardinal axes of the head (axial, sagittal, coronal). However, this was often limited by the size and shape of the craniotomy. The transducer was swept unidirectionally at a slow, consistent speed through the craniotomy. To enable tracking, the ``Ultrasound Navigation Adapter Array'' together with the ``Ultrasound Navigation Adapter Base - BK N13C5'' (Brainlab AG, Munich, Germany) were attached to the iUS probe. This enabled the reconstruction of a 3D volume from the tracked 2D sweeps using the ``Ultrasound'' module within the ``Elements'' software platform on the ``Curve'' hardware system.

    \subsubsection{Data pre-processing.}  Common pre-processing to the same voxel resolution ($0.5\times0.5\times0.5$mm) and spatial dimension ($256\times256\times256$) is performed on all the images. Because the iUS volumes were reconstructed using a calibrated neuronavigation tracking system, they are already placed in the same approximate image space as the preoperative MRI. The remaining misalignment is therefore primarily due to brain shift deformations.

    \subsubsection{Annotation protocol.} The reference annotations were generated through a semi-automatic procedure followed by expert manual refinement. First, salient anatomical features were automatically detected using 3D-SIFT~\cite{chauvin2020neuroimage} in the iUS volumes acquired before dural opening (not included in this challenge) and after resection. Features identified in the pre-dural images were transferred to the preoperative MR space using affine image registration~\cite{drobny2018registration}, taking advantage of the absence of non-linear or topologically changing deformations and the accurate initialization provided by the navigation system. Two medical imaging experts (RD and WW), each with over five years of experience in brain anatomy, then identified approximately ten corresponding anatomical landmarks for each pair of post-resection iUS and T2-SPACE. Eligible landmarks included deep grooves and corners of sulci, convex points of gyri, and vanishing points of sulci. The two annotators performed the landmark selection alternately, with each proposal reviewed and adjusted by the other until consensus was reached. After completing all annotations, both experts jointly reviewed every landmark and discarded those lacking unanimous agreement. To obtain the ceT1 landmarks, the affine deformation to align T2-SPACE scans with ceT1 scans was computed to NiftyReg~\cite{niftyreg}, and applied to the T2-SPACE landmarks.

    \subsubsection{Potential errors in annotation protocols.} The inter-rater variability between pre-dura and post-resection was found to be equal to $1.89 \pm 0.37$ mm in a previous study~\cite{machado2018non} using a similar annotating technique. This inter-rater variability may be higher as we add an extra step (transferring to the MR data).

    \subsection{Challenge setup}

    The Grand Challenge\footnote{\url{https://grand-challenge.org/}} platform was utilized for the ReMIND2Reg 2025 challenge.  Grand Challenge is a well-established platform for managing automated validation leaderboards. For the validation phase, participants were invited to submit their predictions on the validation set directly on the platform for automatic assessment.

    External data and pre-trained models are allowed. Members of the organizing institutes can participate in the challenge but are not eligible for awards. The winning teams of each edition will receive a prize. The top 5 teams will be invited to present their methods at the MICCAI Learn2Reg workshop. The first (co-)author(s) and senior author of the submitted short paper are qualified as authors for the joint publication. 

    To evaluate the accuracy of predictions made on local machines, predictions on the validation set will be computed and compared with those obtained from participants' machines. Participants will be required to include predictions for all cases in the submitted zip file. Submissions with incomplete predictions will not be evaluated or counted as valid validation submissions. The evaluation code is available on GitHub\footnote{\url{https://github.com/ReubenDo/ReMIND2Reg}}. 
    
    Adhering to best practice guidelines for challenge organization~\cite{maier2020bias}, the test set is kept private to mitigate the risk of cheating. Participants are required to containerize their methods using Docker, following the provided guidelines. Each team is permitted a single submission, and the Docker containers are executed on a cluster for evaluation. As a result, no missing results are expected for the test set.

    \subsection{Participation Timeline}  
        
        The challenge is composed of three main stages:
                    
            \begin{enumerate}
                \item  \textbf{Training}: The unregistered pairs of preoperative MRI and post-resection 3D iUS are shared with participants to design and train their methods.
                \item  \textbf{Validation}: The validation data was released to the participants after the training data. The participants were not provided with the ground truth of the validation data, but were allowed to submit multiple times to the online grand-challenge platform. The top-5 ranked participating teams in the validation phase will be invited to prepare their slides for a short oral presentation of their method during the Learn2Reg challenge at MICCAI 2025. 
                \item  \textbf{Testing/Ranking}: After the participants share their containerized methods for each MR sequence, they will be evaluated and ranked on private test data, which will not be made available. The final top-ranked participating teams will be announced at the 2025 MICCAI Annual Meeting. 
            \end{enumerate}

    \subsection{Evaluation metrics and ranking procedure}
    
    The evaluation of registration performance considers three key aspects: accuracy, robustness, and computational efficiency. We report the following metrics:
    
    \begin{itemize}
    \item \textbf{TRE}: Target registration error, defined as the Euclidean distance between corresponding landmarks in the fixed image and the warped moving image.
    \item \textbf{TRE30}: To assess robustness, TRE30 reports the 30\textsuperscript{th} percentile of the largest landmark distances, highlighting worst-case performance with respect to TRE.
    \item \textbf{Runtime}: The mean execution time (including image I/O), computed over the test sets on an A100 GPU (40gb memory).
    \end{itemize}
    
    In line with previous editions of the Learn2Reg Challenge~\cite{hering2023tmi} and the Medical Decathlon~\cite{antonelli2022medical}, method ranking is based on the Wilcoxon signed-rank test.
    Unlike other Learn2Reg tasks~\cite{chen2025lumirchallengepathwayfoundational}, the number of landmarks varies across cases in this dataset from 6 to 18 for the validation set and 4 to 16 for the test set. To account for potential bias due to an imbalanced number of landmarks per case, we perform bootstrap sampling ($B = 100$) using the minimum number of landmarks available across all sets ($L=6$ for the validation set, $L=4$ for the test set). For each bootstrap sample, $L$ landmarks per case are randomly selected without replacement, and the following ranking procedure is applied:
    
    \begin{enumerate}
    \item For \textbf{TRE}, all methods are compared pairwise using the paired Wilcoxon signed-rank test (one-sided alternative hypothesis, per-comparison $\alpha = 0.05$, no multiple-comparison adjustment). Each method’s score is the number of “wins” across comparisons, normalized to values between 0.1 and 1. Ties are possible.
    \item For \textbf{TRE30}, the same procedure is applied but using the unpaired Wilcoxon rank-sum test \sloppy(Mann-Whitney U), as the landmark sets differ between methods for this statistic.
    \item The \textbf{bootstrap task rank score} is computed as the geometric mean of the individual metric rank scores.
    \end{enumerate}
    
    Finally, the final task rank score is obtained by averaging the $B$ bootstrap task rank scores.

        \section{Discussion}
In this paper, we outlined the design of the ReMIND2Reg 2025 challenge to benchmark registration methods designed for preoperative MRI and 3D post-resection intra-operative ultrasound. We are actively working on increasing the number of subjects in this cohort to provide the community with a large dataset and to facilitate the future development of tools for computer-assisted surgery. Future ReMIND2Reg challenges will include data from more institutions and more variability in MRI data.

    \section*{Acknowledgments}
    This work was supported by a Marie Skłodowska-Curie grant No 101154248 (project: SafeREG). We would like to acknowledge support received from the NIH through grant R01EB032387. Additionally, we are thankful for the philanthropic support from the Jennifer Oppenheimer Cancer Research Initiative and Karen and Brian McMahon. 
    We would also like to thank the Medical Image Computing and Computer Assisted Intervention (MICCAI) Society for their invaluable support of this challenge.

    \bibliographystyle{ieeetr}
    \bibliography{bibliography.bib}
    \newpage
    \appendix
\end{document}